\documentclass[preprint,12pt]{elsarticle}

\usepackage{amssymb}
\usepackage{amsmath}

\journal{Journal of Manufacturing System}

\begin{document}

\begin{frontmatter}



\title{AI Agents and Agentic AI–Navigating a Plethora of Concepts for Future Manufacturing} 


\author{Yinwang Ren} 
\author{Yangyang Liu}
\author{Tang Ji}
\author{Xun Xu\corref{cor1}}
\cortext[cor1]{Corresponding author: Xun Xu (x.xu@auckland.ac.nz)}

\affiliation{organization={Department of Mechanical and Mechatronics Engineering, Faculty of Engineering and Design, University of Auckland},
            addressline={20 Symonds Street}, 
            city={Auckland},
            postcode={1010}, 
            country={New Zealand}}

\begin{abstract}
AI agents are autonomous systems designed to perceive, reason, and act within dynamic environments. With the rapid advancements in generative AI (GenAI), large language models (LLMs) and multimodal large language models (MLLMs) have significantly improved AI agents' capabilities in semantic comprehension, complex reasoning, and autonomous decision-making. At the same time, the rise of Agentic AI highlights adaptability and goal-directed autonomy in dynamic and complex environments. LLMs-based AI Agents (LLM-Agents), MLLMs-based AI Agents (MLLM-Agents), and Agentic AI contribute to expanding AI’s capabilities in information processing, environmental perception, and autonomous decision-making, opening new avenues for smart manufacturing. However, the definitions, capability boundaries, and practical applications of these emerging AI paradigms in smart manufacturing remain unclear. To address this gap, this study systematically reviews the evolution of AI and AI agent technologies, examines the core concepts and technological advancements of LLM-Agents, MLLM-Agents, and Agentic AI, and explores their potential applications in and integration into manufacturing, along with the potential challenges they may face.
\end{abstract}




\end{frontmatter}




\section{Introduction}

As a complex and data-intensive domain, manufacturing faces increasing challenges due to the increasing demand for customization, shorter product life cycles, and intense global competition \cite{RN16, RN45}.  Traditional automated systems, reliant on fixed rules, struggle to adapt to evolving customer needs. Although advanced robotics and classic machine learning have improved productivity, they remain constrained by predefined features and limited datasets \cite{RN17}, making them ineffective in handling unstructured data or novel scenarios. In addition, manufacturing requires real-time response, precise control, and the integration of continuous and discrete decision-making \cite{RN18}. These challenges highlight the need for more flexible, adaptive, and intelligent AI-driven solutions.

The rapid rise of Generative AI (GenAI) has reshaped multiple industries, from content creation and software development to scientific research and business automation \cite{RN52, RN60, RN61}. Large Language Models (LLMs), such as ChatGPT, exhibit unprecedented capabilities in natural language comprehension, autonomous reasoning, and cross-domain knowledge synthesis \cite{RN62, RN19}. Concurrently, Multimodal Large Language Models (MLLMs) extend these capabilities beyond text by integrating visual, sensor, and structured data, thereby enabling more sophisticated, context-aware decision-making \cite{RN63, RN64}.

With these advances in GenAI, AI agents have regained attention as systems capable of perception, reasoning, and action \cite{RN90}. Recent work has explored how LLMs and MLLMs can be integrated into AI agents (LLM-Agents, MLLM-Agents) to expand their adaptability and decision-making potential \cite{RN46, RN48}. At the same time, the emerging Agentic AI paradigm represents a transition to self-directed, adaptive, and goal-driven intelligence, enabling autonomous optimization and strategic decision-making in dynamic environments \cite{RN9, RN27}, a typical scenario of manufacturing processes.

LLM-Agents, MLLM-Agents, and Agentic AI are believed to have contributed to expanding AI capabilities in different ways, improving processing information, environmental awareness, and autonomous decision-making. The advancement of these technologies drives the evolution of AI agents and opens new possibilities for future manufacturing systems. However, despite their transformative potential, the definitions, capability boundaries, application contexts, and interconnections of emerging AI paradigms in manufacturing still require further clarification.

This paper first systematically analyses the technological evolution of AI and AI agents, examining the core concepts, technological advancements, and capability enhancements of LLM-Agents, MLLM-Agents, and Agentic AI. This is followed by how these advancements can be deeply integrated into manufacturing. Finally, potential challenges are assessed.

\section{The Development of AI and Agents}

GenAI-enabled AI agents can potentially accelerate the intelligent transformation of the manufacturing industry. However, to gain a deeper understanding of the practical applications of these technologies in manufacturing, it is necessary to explore various implementation methods and theoretical paradigms based on the fundamental theories of artificial intelligence. This section will review the historical development of artificial intelligence and the field of agents. By outlining these foundational theories and technological evolutions, we hope to provide a solid theoretical basis for subsequent discussions on the advantages and challenges of AI agents in manufacturing applications.

\subsection{Evolutionary Path of AI Technologies}
\begin{figure}[h]
	\vspace*{4pt}
	\centering
	\includegraphics[width=\textwidth]{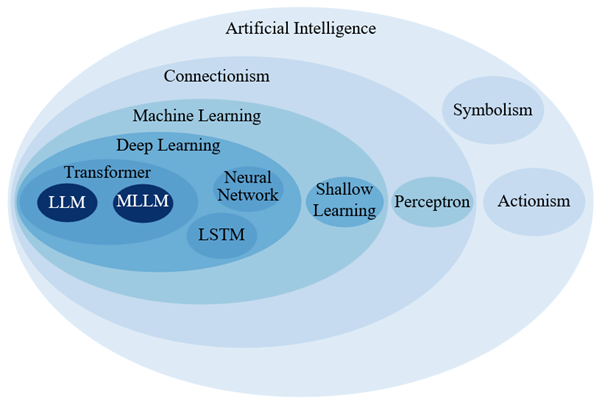}
	\caption{Relationship from AI to LLMs and MLLMs}
	\label{fig.1}
\end{figure}

AI was first explicitly proposed by John McCarthy in 1956 and was defined as "the science and engineering of making intelligent machines" \cite{RN28}. AI aims to enable computers to possess human-like intelligence, encompassing perception, reasoning, and decision-making, thereby enhancing their ability to adapt autonomously in complex environments \cite{RN29, RN30}. There are many different implementation methods or technologies (their relationship is shown in Figure \ref{fig.1}).

Research in AI primarily falls into three paradigms: Symbolism, Connectionism, and Actionism \cite{RN31}. Connectionism has emerged as the mainstream approach in recent years, heavily relying on machine learning (ML) as its core methodology. ML automatically learns patterns and knowledge from data and is mainly categorized into Shallow Learning and Deep Learning (DL) \cite{RN32}. Shallow Learning methods (e.g., logistic regression, support vector machines) rely on manually designed features and simple model architectures, making them suitable for basic tasks. In contrast, DL employs multi-layer neural networks (such as convolutional neural networks \cite{RN33}, Transformers \cite{RN34}, and LSTM \cite{RN35}) to significantly enhance models' capability to represent complex nonlinear relationships. The capacity of deep learning to approximate complex, nonlinear functions in high-dimensional spaces renders it particularly effective for modelling the heterogeneous, data-intensive, and multivariable processes inherent in modern manufacturing systems \cite{RN76}. ML methods, particularly DL, have been widely adopted in manufacturing, supporting tasks such as predictive maintenance \cite{RN75}, process optimization \cite{RN78}, and human-robot collaboration \cite{RN77}.

The introduction of the Transformer architecture in 2017 ushered in the era of large-scale pre-trained models in deep learning, significantly advancing performance in natural language processing tasks. With advancements in computational power and the extensive application of big data, a series of Transformer-based large-scale language models have emerged, including the GPT series \cite{RN19}, Llama \cite{RN36}, and Qwen  \cite{RN72}. 

LLMs exhibit powerful capabilities in contextual understanding, instruction following, and step-by-step reasoning \cite{RN37}. However, their primary limitation lies in the confinement to unimodal linguistic processing, which significantly hinders cross-modal cognitive capabilities in industrial scenarios. This deficiency becomes particularly pronounced in manufacturing systems where multi-source heterogeneous data streams (text, image, 3D prints, formula and so on) require synergistic interpretation. MLLMs have been developed to overcome this limitation. Representative models include GPT-4V \cite{RN74}, LLaVA \cite{RN38}, and mPLUG-Owl2 \cite{RN39}. 

MLLMs demonstrate strong generalization capabilities and complex reasoning beyond what single-modal models can achieve \cite{RN73}. They learn cross-modal semantic representations through training on massive multimodal datasets while reducing reliance on task-specific annotations. Moreover, with minimal fine-tuning, MLLMs can adapt to new tasks via prompting or few-shot learning, thereby supporting dynamic reasoning across modalities. 

\subsection{Evolutionary Path of Agent Technologies}

As early as the 1950s, Alan Turing expanded the concept of "intelligence" to artificial entities and proposed the renowned Turing Test. These artificial intelligence entities are generally referred to as agents  \cite{RN40}. Indeed, agents and AI are often inseparable in their conception, functions and applications. Early agents were primarily designed for solving specific tasks, such as expert systems and rule-based reasoning systems \cite{RN41}. These systems simulated expert decision-making processes through "if-then" rule chains but lacked learning capabilities and adaptability to complex scenarios. 

With the advancement of computational power and knowledge engineering, agent technology evolved into multi-agent systems (MAS), enabling multiple agents to collaborate in accomplishing complex tasks, such as traffic control, robotic swarms, and financial trading \cite{RN41}. In the 21st century, the decision-making capabilities of AI-enabled agents have been significantly enhanced, allowing them to learn and adapt autonomously in dynamic environments, as exemplified by AlphaGo \cite{RN42}. 

Traditional agent research for manufacturing has primarily centered on algorithm design and training strategies, often overlooking core capabilities such as knowledge retention, long-term planning, generalization, and efficient interaction \cite{RN40}. Strengthening these fundamental abilities is essential for the advancement of intelligent agents, particularly in the domain of manufacturing. The unique pretraining architecture and emergent abilities of LLMs and MLLMs not only compensate for the deficiencies of early agents in knowledge retention, long-term planning, and dynamic adaptation but also facilitate the transition from rule-based tools to partner-like intelligent systems with autonomous cognition, real-time interaction, and multimodal collaboration through instruction fine-tuning and multimodal alignment techniques.

\section{From GenAI-enabled AI Agents to Agentic AI}

The evolution of AI agents has been driven by advancements in GenAI, enabling increasingly autonomous, adaptive, and multimodal capabilities. This section examines the transition from GenAI-enabled AI Agents to Agentic AI, tracing how large language models LLMs, MLLMs, and agentic architectures contribute to this progression.

\subsection{LLM-Agents}

In recent years, GenAI has significantly advanced AI agent development, with LLMs playing a central role. Their strong language comprehension, reasoning, and decision-making abilities \cite{RN12}, enable LLM-based agents to handle complex task planning, problem-solving, and human-machine collaboration.

\begin{figure}[h]
	\vspace*{4pt}
	\centering
	\includegraphics[width=\textwidth]{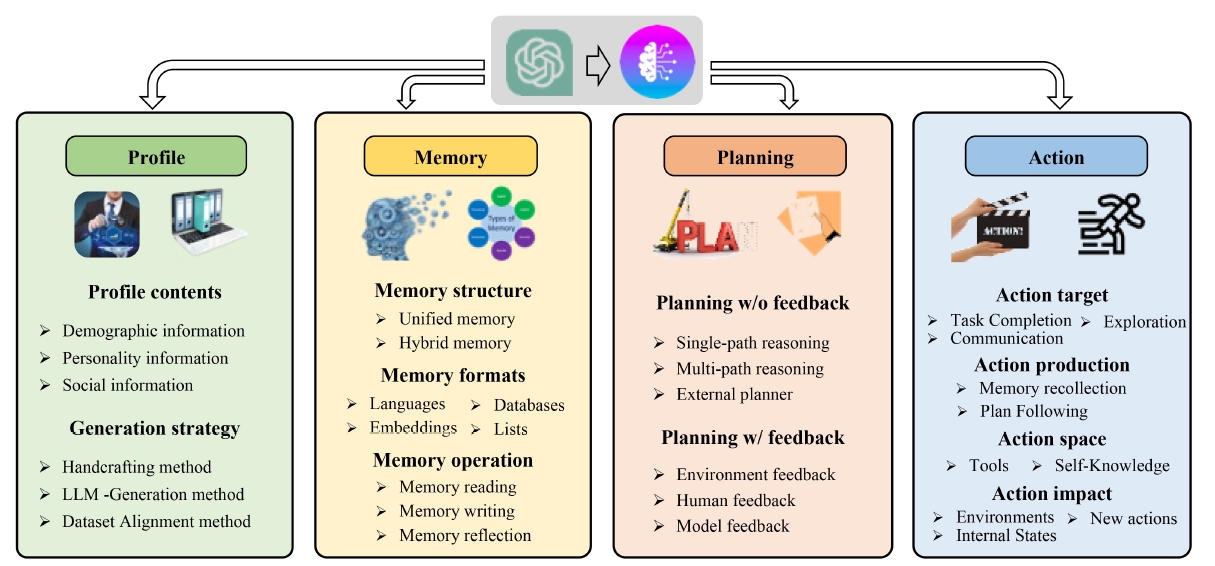}
	\caption{Different components and its functions of LLM-Agents \cite{RN12}}
	\label{fig.2}
\end{figure}

The architecture of LLM-Agents typically consists of four core components \cite{RN12} (as shown in Figure \ref{fig.2}). The Profiling Module defines the agent’s identity, role, and behavioural constraints. The Memory Module stores and retrieves past interactions, improving contextual awareness for decision-making. The Planning Module decomposes complex tasks into structured steps, ensuring adaptability across various applications. Lastly, the Action Module executes decisions using external tools or internal knowledge synthesis. Together, these components enable LLM-Agents to operate autonomously within predefined constraints, leveraging both static knowledge and dynamic learning mechanisms.

Compared to traditional AI models, LLM-Agents offer notable advantages. First, they possess extensive domain knowledge and strong reasoning abilities, enabling them to process complex textual inputs efficiently. Second, their generalization capabilities allow them to perform tasks without specific training, leveraging zero-shot or few-shot learning. Additionally, their natural language understanding facilitates more intuitive and context-aware interactions. However, despite their strengths in text processing, LLM-Agents primarily rely on language inputs, limiting their ability to perceive and process non-textual information effectively. Furthermore, their autonomy remains constrained, as they often require predefined tasks or external instructions to guide decision-making and execution.

\subsection{MLLM-Agents}
To overcome the limitations of LLM-Agents, researchers have introduced MLLM-Agents that can process text, images, audio, video, and structured data. By integrating multiple data types, these agents gain a deeper understanding of their environment, enabling more precise perception, reasoning, and decision-making. This makes them particularly useful in fields such as robotics, autonomous systems, and human-computer interaction.

Unlike LLM-Agents, which primarily rely on textual inputs, MLLM-Agents adopt a more sophisticated architecture. Their Multimodal Perception Module first collects and integrates various data types. Subsequently, the Fusion and Reasoning Module synthesizes these inputs, forming a comprehensive environmental representation. The Decision and Planning Module then uses this information to formulate strategies, and the Action and Execution Module carries out tasks across different data formats, allowing seamless interaction with complex surroundings.

One key advantage of MLLM-Agents is their adaptability, as they can process both language and sensory data, leading to more informed decisions in dynamic environments. They are also more resilient in uncertain and data-rich conditions. However, these benefits come with challenges. Handling multiple data streams requires significant computational power, and integrating different types of information can introduce inconsistencies that affect accuracy. Scaling these systems for large-scale applications remains another obstacle. Despite these hurdles, MLLM-Agents represent a major step forward in bridging language-based intelligence with real-world perception.

\subsection{Agentic AI }

Agentic AI is emerging as a pivotal paradigm in artificial intelligence, denoting autonomous systems capable of independently pursuing complex objectives with minimal human oversight in dynamic and uncertain environments \cite{RN9}. Recognizing its transformative potential, Gartner has identified Agentic AI as the top strategic technology trend for 2025 \cite{RN79}.

To better understand agentic AI, let us talk about the concept of agenticness. Agenticness is defined as the degree to which a system can adaptably achieve complex goals in dynamic environments with limited direct supervision \cite{RN43}. It encompasses four key dimensions:
\begin{itemize}
  \item \textbf{Goal Complexity:} The range and difficulty of tasks the system can accomplish, considering reliability, speed, and safety.
  \item \textbf{Environmental Complexity:} The system's ability to operate across diverse, multi-stakeholder, or long-horizon contexts.
  \item \textbf{Adaptability:} The extent to which the system can respond to novel or unexpected circumstances.
  \item \textbf{Independent Execution:} The system’s capability to autonomously achieve objectives with minimal human intervention.
\end{itemize}

OpenAI defines systems with high agenticness as "Agentic AI systems," emphasizing that agentic is a gradual spectrum rather than a fixed classification \cite{RN43}. Thus, rather than a strict dichotomy between "agentic" and "non-agentic," AI systems exhibit varying degrees of agenticness, continuously evolving as their capabilities in autonomy, adaptability, and goal-directed reasoning advance. As these capabilities reach a sufficiently high threshold, AI agents naturally transition into Agentic AI systems.

\begin{figure}[h]
	\vspace*{4pt}
	\centering
	\includegraphics[width=\textwidth]{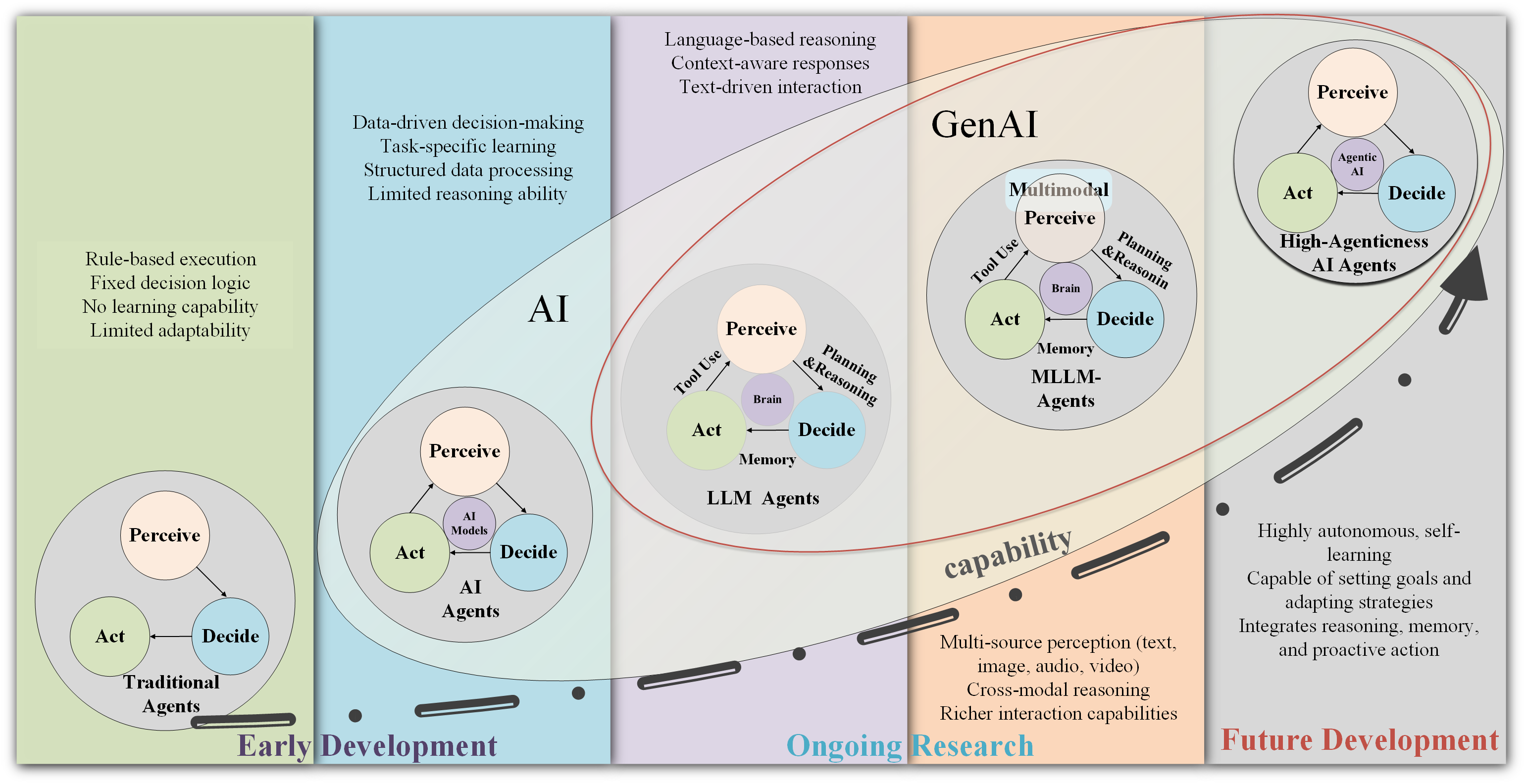}
	\caption{Evolutionary Path of AI Agent Technologies}
	\label{fig.3}
\end{figure}

Figure\ref{fig.3} shows the evolution of agent technology that has progressed through five key stages from early rule-based expert systems to AI-powered decision-making, followed by LLM-based agents, and now, MLLM-based agents with richer cognitive capabilities. The next phase of development, agentic AI, represents a paradigm shift where AI systems exhibit high levels of autonomy, adaptive learning, and proactive decision-making.

\section{GenAI-enabled AI Agents in Manufacturing}
The integration of GenAI-enabled AI Agents in manufacturing marks a fundamental transformation from conventional rule-based automation to intelligent, self-optimizing systems capable of knowledge retrieval, multimodal reasoning, self-learning, and autonomous decision-making. Unlike traditional AI approaches that rely heavily on structured data and predefined models, GenAI-enabled AI Agents bring semantic understanding, context-awareness, and adaptive reasoning to manufacturing processes. This section explores the progressive capabilities of GenAI-enabled AI Agents, including Semantic Retrieval, Cognitive Perception, Self-Learning Optimization, and Autonomous Decision-Making, as well as their transformative impact on the manufacturing industry.

\subsection{Knowledge-Enhanced Semantic Retrieval and Automated Documentation}
Manufacturing enterprises operate in highly complex environments where critical knowledge is distributed across heterogeneous IT and OT systems, including Enterprise Resource Planning (ERP), Manufacturing Execution Systems (MES), Product Lifecycle Management (PLM), and Supervisory Control and Data Acquisition (SCADA). In addition to structured data, unstructured sources such as maintenance logs, operation manuals, regulatory guidelines, and production reports contain vital decision-making information but remain fragmented and difficult to integrate.

GenAI-enabled AI Agents, leveraging Retrieval-Augmented Generation (RAG) \cite{RN67} and knowledge graphs \cite{RN68}, bridge these gaps by facilitating semantic knowledge retrieval and automated documentation synthesis. Unlike traditional AI approaches that rely on keyword-based search and structured queries, these agents employ context-aware natural language processing to dynamically extract and synthesize manufacturing data and knowledge. By harmonizing structured and unstructured data, engineers and operators can access precise, contextually relevant insights through natural language interactions, significantly improving operational efficiency and knowledge accessibility.

Recent research demonstrates the effectiveness of such systems in real-world applications. Lin et al. \cite{RN49} proposed an Intelligent Manufacturing Virtual Assistant (IMVA) for the semiconductor industry, leveraging LLM Agents to integrate diverse manufacturing systems, enabling natural language interaction, real-time data retrieval, and automated report generation for improved efficiency and fault diagnosis. Jeon et al. \cite{RN47} introduced ChatCNC, a GenAI-enabled AI Agent framework with real-time RAG, enabling natural language queries on CNC data and IIoT sensor readings, enhancing decision-making while reducing reliance on structured queries.

By moving beyond rigid keyword-based queries and enabling cross-system semantic integration, these systems enhance retrieval precision, improve interpretability, and optimize decision-making efficiency through natural language interaction and automated knowledge synthesis.

\subsection{Multimodal Cognitive Perception and Contextual Reasoning}

Modern manufacturing environments generate multimodal data spanning structured system outputs (ERP, MES, SCADA, PLC), unstructured text (maintenance logs, technical manuals, quality reports, operator notes), and real-time sensor and machine vision inputs (images, video, thermal scans, vibration, acoustic signals, temperature, pressure). While some traditional AI models incorporate multimodal fusion, they primarily rely on feature-based learning and predefined patterns, limiting their ability to perform complex reasoning and integrate domain-specific knowledge dynamically. As a result, conventional approaches excel at detecting correlations but struggle with causal inference, adaptive diagnostics, and real-time problem-solving, which are critical elements for predictive maintenance, fault diagnosis, and quality control.

GenAI-enabled AI Agents overcome these limitations by integrating MLLMs with knowledge retrieval, enabling context-aware perception and reasoning. By fusing real-time sensor data, machine vision insights, and IT-OT system outputs with retrieved domain knowledge from maintenance logs and technical manuals, they provide accurate diagnostics and proactive recommendations. Unlike traditional AI, which operates on isolated datasets, these agents perform cross-modal correlation, knowledge retrieval, and automated decision support, advancing from anomaly detection to explainable diagnostics and prescriptive recommendations. Heredia Álvaro et al. \cite{RN66} developed an RAG system leveraging MLLMs to diagnose defects and propose solutions in ceramic tile production. By integrating bibliographic sources with real-time sensor data and visual defect analysis, the system demonstrates how GenAI-enabled AI Agents can synthesize multimodal domain knowledge and enhance manufacturing diagnostics.

Beyond quality control, these capabilities hold significant potential for other key areas of manufacturing. In predictive maintenance, AI agents could analyse vibration and thermal anomalies in industrial equipment, retrieve maintenance histories and manufacturer specifications, and infer the likelihood of failure, recommending proactive interventions. While these capabilities are still emerging in manufacturing, advancements in MLLMs for medical diagnostics \cite{RN70} and robotics \cite{RN71} indicate strong potential for manufacturing applications

\subsection{Adaptive Learning and Evolutionary Optimization}

Modern manufacturing operates in dynamic, uncertain environments where fluctuations in demand, resource availability, machine performance, and supply chain conditions require continuous adaptation.

GenAI-enabled AI Agents integrate multimodal cognitive perception with adaptive learning and evolutionary optimization, enabling systems to not only interpret and reason across diverse industrial data but also continuously refine optimization strategies based on real-time feedback. Unlike traditional AI, which relies on fixed-rule optimizations, these agents dynamically learn from sensor data, machine vision insights, and historical operational trends, allowing them to adapt strategies in response to evolving manufacturing conditions.

While practical implementations remain in the early stages, we anticipate that such AI-driven frameworks could transform key manufacturing processes:
\begin{itemize}
  \item \textbf{Production Scheduling:} GenAI-enabled AI Agents dynamically adjust machine allocations, prioritize orders, and optimize workforce distribution by correlating real-time IT-OT system outputs with historical production patterns and knowledge-based recommendations.
  
  \item \textbf{Process Optimization:} By integrating sensor data and quality reports with domain knowledge from technical manuals and maintenance logs, GenAI-enabled AI Agents continuously refine manufacturing parameters, energy efficiency strategies, and defect mitigation processes through self-optimizing mechanisms.
  
  \item \textbf{Supply Chain Resilience:} GenAI-enabled AI Agents synchronize procurement, logistics, and inventory management by analysing supplier reliability, transportation delays, and demand fluctuations, ensuring proactive adjustments to minimize risks and optimize cost-effectiveness.
\end{itemize}

By unifying multimodal perception, Contextual reasoning, and adaptive learning, GenAI-enabled AI Agents move beyond static optimization, fostering self-evolving, autonomous decision-making frameworks that enhance manufacturing agility and resilience.

Integrating GenAI-enabled AI agents into manufacturing has significantly enhanced knowledge retrieval, multimodal perception and contextual reasoning, adaptive learning, and optimization. These advancements enhance manufacturing agility and resilience by continuously refining decision-making strategies in response to evolving conditions. However, these agents remain task-oriented, optimizing predefined objectives rather than autonomously shaping manufacturing strategies.

As manufacturing systems grow in complexity, AI must evolve beyond adaptive optimization toward autonomous, goal-driven decision-making. This transition marks the emergence of Agentic AI, where AI systems take on greater responsibility in defining, managing, and executing manufacturing processes. This transformation is explored in the following chapter.

\section{Agentic AI for Future Manufacturing}
Agentic AI’s emphasis on autonomy, adaptability, and goal-driven decision-making resonates strongly with the strategic imperatives of contemporary manufacturing. Unlike existing AI systems that optimize predefined tasks, Agentic AI introduces a new paradigm in which manufacturing systems can autonomously define, refine, and execute goals in dynamic environments with minimal human intervention.

Although Agentic AI is still evolving, it signals a move from reactive task optimization to proactive system-level intelligence, emphasizing autonomy, adaptability, system-wide coordination, and continuous learning. The following sections explore these transformations and their implications for the future of manufacturing. 

\subsection{From Task Execution to Goal-Driven Optimization}
Current AI optimizes well-defined objectives such as scheduling and quality inspection but lacks the ability to autonomously redefine optimization priorities in response to external changes.

Agentic AI introduces self-directed goal formulation, allowing systems to dynamically adjust production objectives based on market conditions, supply chain variability, and real-time operational data. Instead of optimizing fixed schedules, Agentic AI can autonomously determine how to maximize throughput, balance energy efficiency, and optimize resource allocation under shifting constraints. This shift transforms manufacturing from reactive automation to proactive intelligence, enabling greater adaptability in volatile industrial environments.

\subsection{From Rule-Based Control to Adaptive Planning}
Traditional AI in manufacturing follows static rules, requiring manual intervention to adjust decision-making models when conditions change. This limits its ability to operate effectively in unpredictable and rapidly evolving environments.

Agentic AI, powered by reinforcement learning, multimodal AI, and real-time analytics, enables adaptive, self-optimizing production strategies. For instance, if a supply chain disruption occurs, an Agentic AI system can autonomously modify production workflows, identify alternative materials, and reconfigure supply logistics without human oversight. By enabling real-time, data-driven adaptability, Agentic AI moves beyond deterministic control systems to fluid, continuously optimizing decision-making frameworks.

\subsection{From Localized Optimization to System-Level Orchestration}
Most AI applications in manufacturing are compartmentalized, optimizing individual components such as warehouse automation, equipment maintenance, or production line scheduling. However, these fragmented solutions require human coordination across subsystems, limiting scalability and adaptability.

Agentic AI orchestrates intelligence across production, logistics, and enterprise management, ensuring cross-functional optimization at a system-level. It enables autonomous synchronization of scheduling, inventory management, and transportation logistics, reducing inefficiencies caused by siloed decision-making. This capability is crucial for high-mix, low-volume manufacturing environments, where dynamic coordination is essential to balance cost, efficiency, and responsiveness.

\subsection{From Static Execution to Continuous Learning and Evolution}
Conventional AI in manufacturing requires periodic retraining to remain relevant, limiting its ability to adapt dynamically to changing conditions. This static approach prevents AI systems from evolving autonomously.

Agentic AI integrates self-supervised learning and reinforcement learning, enabling manufacturing systems to refine decision-making strategies in real time. Instead of relying on periodic updates, Agentic AI can continuously improve its models, optimizing energy usage, reducing waste, and enhancing predictive maintenance over extended operational cycles. This evolutionary capability unlocks the potential for long-term industrial intelligence, where AI systems become more efficient and resilient over time without human intervention.

Agentic AI in manufacturing advances AI capabilities from fixed-task execution to self-directed optimization, from static control to adaptive decision-making, and from localized automation to system-wide autonomy. By integrating real-time learning mechanisms, Agentic AI has the potential to transform manufacturing ecosystems into self-optimizing, continuously evolving systems. While the full realization of Agentic AI remains an ongoing challenge, its potential to enhance manufacturing resilience, efficiency, and adaptability makes it a compelling avenue for future research.

\section{Challenge}
\subsection{Technology Challenges}

The integration of AI Agents and Agentic AI in manufacturing faces significant challenges due to the heterogeneous nature of industrial knowledge, the complexity of multimodal data, and the necessity for interpretability in AI-driven decision-making.

\subsubsection{Cross-Format Document Parsing in Manufacturing}
Manufacturing knowledge is predominantly stored in multi-source, unstructured documents such as technical manuals and assembly specifications, often generated through diverse formats, such as Word-to-PDF conversions, LaTeX rendering, and scanned image recognition. Existing PDF parsing techniques struggle with text fragmentation, formula distortion, and vector graphic loss, necessitating a robust cross-format document reconstruction framework to ensure semantic consistency and reliability.

\subsubsection{Multimodal Knowledge Extraction and Alignment in Manufacturing}

Manufacturing knowledge is inherently multimodal, encompassing text, mathematical equations (LaTeX, MathML), engineering schematics, and CAD. Effectively extracting implicit process parameters, equipment constraints, and engineering relationships from vast unstructured datasets requires context-aware LLMs integrated with knowledge graphs. A critical challenge lies in achieving fine-grained cross-modal semantic alignment, ensuring that textual descriptions, mathematical derivations, and visual representations are accurately correlated and complement each other within a unified knowledge framework.

\subsubsection{Interpretability and Explainability in Manufacturing}

Interpretability is a critical prerequisite for AI deployment in manufacturing, as decision-making processes must be transparent, verifiable, and aligned with domain-specific constraints. However, current LLMs function as opaque black-box models, lacking the capacity for structured, explainable reasoning, thereby constraining their reliability in high-stakes industrial applications. Overcoming this limitation necessitates the integration of causal inference, physics-informed modelling, and explainable AI (XAI) frameworks, ensuring that AI-driven decisions are traceable, auditable, and actionable within complex manufacturing ecosystems.

\subsection{Workforce and Organizational Resistance}
Implementing next-generation AI demands interdisciplinary collaboration between engineers, data scientists, and operators. Yet, rigid corporate structures, traditional workflows, and limited AI literacy create resistance to change. Many employees lack the necessary technical expertise to interpret AI-generated insights or interact effectively with autonomous systems. Additionally, resistance to change can slow deployment, particularly in environments where manual intervention has historically dominated production processes. Addressing these challenges requires AI-focused workforce training programs, skill augmentation initiatives, and leadership-driven organizational adaptation strategies to bridge the gap between human expertise and AI-driven decision-making.

\subsection{Accountability and ROI Concerns}
As AI Agents take on greater autonomy in manufacturing decision-making, ensuring accountability and governance becomes critical. In safety-critical operations such as quality control, predictive maintenance, and production optimization, manufacturers must define clear responsibility frameworks to audit AI-driven decisions and ensure regulatory compliance.

Additionally, while AI is often associated with efficiency gains and cost reduction, quantifying its return on investment (ROI) remains challenging. In many cases, the benefits of AI-driven systems may not be immediately measurable, making it difficult for businesses to justify large-scale deployment.

\section{Conclusion}
The paper explored the evolution of AI and agent-based systems, focusing on key research trends in LLM-Agents and MLLM-Agents empowered by large models, and the emerging paradigm of Agentic AI. Their development, conceptual foundations, and distinctive characteristics are explained. Their progression toward more autonomous, adaptive, and goal-driven AI systems is of interest.

These AI agents empower smart manufacturing by driving the transition from rule-based automation to intelligent autonomy. Their capabilities in knowledge integration, real-time decision-making, and multimodal perception can enhance manufacturing efficiency, flexibility, and adaptability.  Challenges remain, including data infrastructure, workforce adaptation, and AI accountability, which must be addressed for broader adoption. 

The paper aims to clarify a plethora of concepts and provide a structured perspective on their implications for future manufacturing research. By outlining their technological trajectories and challenges, we aim to stimulate discussions and encourage further investigation into scalable, interpretable, and industrially viable AI agent frameworks that will drive the next generation of smart manufacturing systems.

\section*{Declaration of Competing Interest}
The authors declare that they have no known competing financial interests or personal relationships that could have appeared to influence the work reported in this paper.


\bibliography{cas-refs} 

\begin{thebibliography}{10}
\expandafter\ifx\csname url\endcsname\relax
  \def\url#1{\texttt{#1}}\fi
\expandafter\ifx\csname urlprefix\endcsname\relax\def\urlprefix{URL }\fi
\expandafter\ifx\csname href\endcsname\relax
  \def\href#1#2{#2} \def\path#1{#1}\fi

\bibitem{RN16}
A.~Kusiak, Smart manufacturing, International Journal of Production Research
  (2018).
\newblock \href {https://doi.org/10.1080/00207543.2017.1351644}
  {\path{doi:10.1080/00207543.2017.1351644}}.

\bibitem{RN45}
P.~Zheng, H.~wang, Z.~Sang, R.~Y. Zhong, Y.~Liu, C.~Liu, K.~Mubarok, S.~Yu,
  X.~Xu, Smart manufacturing systems for industry 4.0: Conceptual framework,
  scenarios, and future perspectives, Frontiers of Mechanical Engineering
  13~(2) (2018) 137--150.
\newblock \href {https://doi.org/10.1007/s11465-018-0499-5}
  {\path{doi:10.1007/s11465-018-0499-5}}.

\bibitem{RN17}
J.~Wang, Y.~Ma, L.~Zhang, R.~X. Gao, D.~Wu, Deep learning for smart
  manufacturing: Methods and applications, Journal of Manufacturing Systems 48
  (2018) 144--156.
\newblock \href {https://doi.org/10.1016/j.jmsy.2018.01.003}
  {\path{doi:10.1016/j.jmsy.2018.01.003}}.

\bibitem{RN18}
F.~Tao, H.~Zhang, A.~Liu, A.~Y.~C. Nee, Digital twin in industry:
  State-of-the-art, IEEE Transactions on Industrial Informatics 15~(4) (2019)
  2405--2415.
\newblock \href {https://doi.org/10.1109/tii.2018.2873186}
  {\path{doi:10.1109/tii.2018.2873186}}.

\bibitem{RN52}
J.~Sauvola, S.~Tarkoma, M.~Klemettinen, J.~Riekki, D.~Doermann, Future of
  software development with generative ai, Automated Software Engineering
  31~(1) (2024) 26.
\newblock \href {https://doi.org/10.1007/s10515-024-00426-z}
  {\path{doi:10.1007/s10515-024-00426-z}}.

\bibitem{RN60}
C.~Stokel-Walker, R.~van Noorden, What chatgpt and generative ai mean for
  science, Nature 614 (2023) 214--216.
\newblock \href {https://doi.org/10.1038/d41586-023-00340-6}
  {\path{doi:10.1038/d41586-023-00340-6}}.

\bibitem{RN61}
Z.~Epstein, A.~Hertzmann, C.~the Investigators~of Human, M.~Akten, H.~Farid,
  J.~Fjeld, M.~R. Frank, M.~Groh, L.~Herman, N.~Leach, R.~Mahari, A.~S.
  Pentland, O.~Russakovsky, H.~Schroeder, A.~Smith, Art and the science of
  generative ai, Science 380~(6650) (2023) 1110--1111.
\newblock \href {https://doi.org/10.1126/science.adh4451}
  {\path{doi:10.1126/science.adh4451}}.

\bibitem{RN62}
T.~Wang, J.~Fan, P.~Zheng, An llm-based vision and language cobot navigation
  approach for human-centric smart manufacturing, Journal of Manufacturing
  Systems 75 (2024) 299--305.
\newblock \href {https://doi.org/10.1016/j.jmsy.2024.04.020}
  {\path{doi:10.1016/j.jmsy.2024.04.020}}.

\bibitem{RN19}
T.~B. Brown, Language models are few-shot learners, in: In Proceedings of the
  34th International Conference on Neural Information Processing System, 2020.

\bibitem{RN63}
S.~Yin, C.~Fu, S.~Zhao, K.~Li, X.~Sun, T.~Xu, E.~Chen, A survey on multimodal
  large language models, National Science Review 11~(12) (2024) nwae403.
\newblock \href {https://doi.org/10.1093/nsr/nwae403}
  {\path{doi:10.1093/nsr/nwae403}}.

\bibitem{RN64}
W.~Yu, J.~Lv, W.~Zhuang, X.~Pan, S.~Wen, J.~Bao, X.~Li, Rescheduling
  human-robot collaboration tasks under dynamic disassembly scenarios: An
  mllm-kg collaboratively enabled approach, Journal of Manufacturing Systems 80
  (2025) 20--37.
\newblock \href {https://doi.org/10.1016/j.jmsy.2025.02.015}
  {\path{doi:10.1016/j.jmsy.2025.02.015}}.

\bibitem{RN90}
M.~Wooldridge, N.~R. Jennings, Intelligent agents: theory and practice, The
  Knowledge Engineering Review 10 (1995) 115 -- 152.
\newblock \href {https://doi.org/10.1017/S0269888900008122}
  {\path{doi:10.1017/S0269888900008122}}.

\bibitem{RN46}
J.~Xie, Z.~Chen, R.~Zhang, X.~Wan, G.~Li, Large multimodal agents: A survey,
  ArXiv abs/2402.15116 (2024).

\bibitem{RN48}
J.~Liao, Autoforma: A large language model-based multi-agent for
  computer-automated design, 2024 IEEE International Conference on Systems
  (2024).

\bibitem{RN9}
D.~B. ACHARYA, Agentic ai: Autonomous intelligence for complex goals—a
  comprehensive survey, IEEE Access (2025).
\newblock \href {https://doi.org/10.1109/ACCESS.2025.3532853}
  {\path{doi:10.1109/ACCESS.2025.3532853}}.

\bibitem{RN27}
V.~Shankar, Managing the twin faces of ai: A commentary on “is ai changing
  the world for better or worse?”, Journal of Macromarketing 44~(4) (2024)
  892--899.
\newblock \href {https://doi.org/10.1177/02761467241286483}
  {\path{doi:10.1177/02761467241286483}}.

\bibitem{RN28}
J.~McCarthy, From here to human-level ai, Artificial Intelligence 171~(18)
  (2007) 1174--1182.
\newblock \href {https://doi.org/10.1016/j.artint.2007.10.009}
  {\path{doi:10.1016/j.artint.2007.10.009}}.

\bibitem{RN29}
Y.~LeCun, Y.~Bengio, G.~Hinton, Deep learning, Nature 521~(7553) (2015)
  436--444.
\newblock \href {https://doi.org/10.1038/nature14539}
  {\path{doi:10.1038/nature14539}}.

\bibitem{RN30}
S.~Pouyanfar, S.~Sadiq, Y.~Yan, H.~Tian, Y.~Tao, M.~P. Reyes, M.-L. Shyu, S.-C.
  Chen, S.~S. Iyengar, A survey on deep learning: Algorithms, techniques, and
  applications, ACM Comput. Surv. 51~(5) (2018) Article 92.
\newblock \href {https://doi.org/10.1145/3234150} {\path{doi:10.1145/3234150}}.

\bibitem{RN31}
Q.~Sun, L.~Yang, From independence to interconnection — a review of ai
  technology applied in energy systems, CSEE Journal of Power and Energy
  Systems 5~(1) (2019) 21--34.
\newblock \href {https://doi.org/10.17775/CSEEJPES.2018.00830}
  {\path{doi:10.17775/CSEEJPES.2018.00830}}.

\bibitem{RN32}
S.~Dong, P.~Wang, K.~Abbas, A survey on deep learning and its applications,
  Computer Science Review 40 (2021) 100379.
\newblock \href {https://doi.org/10.1016/j.cosrev.2021.100379}
  {\path{doi:10.1016/j.cosrev.2021.100379}}.

\bibitem{RN33}
D.~E. Rumelhart, G.~E. Hinton, R.~J. Williams, Learning representations by
  back-propagating errors, Nature 323~(6088) (1986) 533--536.
\newblock \href {https://doi.org/10.1038/323533a0}
  {\path{doi:10.1038/323533a0}}.

\bibitem{RN34}
A.~Vaswani, N.~M. Shazeer, N.~Parmar, J.~Uszkoreit, L.~Jones, A.~N. Gomez,
  L.~Kaiser, I.~Polosukhin, Attention is all you need, in: Neural Information
  Processing Systems, 2017.

\bibitem{RN35}
S.~Hochreiter, J.~Schmidhuber, Long short-term memory, Neural Computation 9~(8)
  (1997) 1735--1780.
\newblock \href {https://doi.org/10.1162/neco.1997.9.8.1735}
  {\path{doi:10.1162/neco.1997.9.8.1735}}.

\bibitem{RN76}
A.~Diez-Olivan, J.~Del~Ser, D.~Galar, B.~Sierra, Data fusion and machine
  learning for industrial prognosis: Trends and perspectives towards industry
  4.0, Information Fusion 50 (2019) 92--111.
\newblock \href {https://doi.org/10.1016/j.inffus.2018.10.005}
  {\path{doi:10.1016/j.inffus.2018.10.005}}.

\bibitem{RN75}
O.~Serradilla, E.~Zugasti, J.~Rodriguez, U.~Zurutuza, Deep learning models for
  predictive maintenance: a survey, comparison, challenges and prospects,
  Applied Intelligence 52~(10) (2022) 10934--10964.
\newblock \href {https://doi.org/10.1007/s10489-021-03004-y}
  {\path{doi:10.1007/s10489-021-03004-y}}.

\bibitem{RN78}
Y.~Ren, J.~Dong, J.~He, D.~Zhang, K.~Wu, Z.~Xiong, P.~Zheng, Y.~Sun, S.~Liu, A
  novel six-dimensional digital twin model for data management and its
  application in roll forming, Advanced Engineering Informatics 61 (2024)
  102555.
\newblock \href {https://doi.org/10.1016/j.aei.2024.102555}
  {\path{doi:10.1016/j.aei.2024.102555}}.

\bibitem{RN77}
H.~Liu, L.~Wang, Remote human–robot collaboration: A cyber–physical system
  application for hazard manufacturing environment, Journal of Manufacturing
  Systems 54 (2020) 24--34.
\newblock \href {https://doi.org/10.1016/j.jmsy.2019.11.001}
  {\path{doi:10.1016/j.jmsy.2019.11.001}}.

\bibitem{RN36}
H.~Touvron, T.~Lavril, G.~Izacard, X.~Martinet, M.-A. Lachaux, T.~Lacroix,
  B.~Rozière, N.~Goyal, E.~Hambro, F.~Azhar, A.~Rodriguez, A.~Joulin,
  E.~Grave, G.~Lample, Llama: Open and efficient foundation language models,
  ArXiv abs/2302.13971 (2023).

\bibitem{RN72}
J.~Bai, S.~Bai, Y.~Chu, Z.~Cui, K.~Dang, X.~Deng, Y.~Fan, W.~Ge, Y.~Han,
  F.~Huang, Qwen technical report, arXiv preprint arXiv:2309.16609 (2023).

\bibitem{RN37}
C.~Chen, K.~Zhao, J.~Leng, C.~Liu, J.~Fan, P.~Zheng, Integrating large language
  model and digital twins in the context of industry 5.0: Framework, challenges
  and opportunities, Robotics and Computer-Integrated Manufacturing 94 (2025).
\newblock \href {https://doi.org/10.1016/j.rcim.2025.102982}
  {\path{doi:10.1016/j.rcim.2025.102982}}.

\bibitem{RN74}
Z.~Yang, L.~Li, K.~Lin, J.~Wang, C.-C. Lin, Z.~Liu, L.~Wang, The dawn of lmms:
  Preliminary explorations with gpt-4v (ision), arXiv preprint arXiv:2309.17421
  9~(1) (2023) 1.

\bibitem{RN38}
H.~Liu, C.~Li, Y.~Li, Y.~J. Lee, Improved baselines with visual instruction
  tuning, 2024 IEEE/CVF Conference on Computer Vision and Pattern Recognition
  (CVPR) (2023) 26286--26296.

\bibitem{RN39}
Q.~Ye, H.~Xu, J.~Ye, M.~Yan, A.~Hu, H.~Liu, Q.~Qian, J.~Zhang, F.~Huang,
  J.~Zhou, mplug-owi2: Revolutionizing multi-modal large language model with
  modality collaboration, 2024 IEEE/CVF Conference on Computer Vision and
  Pattern Recognition (CVPR) (2023) 13040--13051.

\bibitem{RN73}
J.~Wu, W.~Gan, Z.~Chen, S.~Wan, P.~S. Yu, Multimodal large language models: A
  survey, in: 2023 IEEE International Conference on Big Data, 2023, pp.
  2247--2256.
\newblock \href {https://doi.org/10.1109/BigData59044.2023.10386743}
  {\path{doi:10.1109/BigData59044.2023.10386743}}.

\bibitem{RN40}
Z.~Xi, W.~Chen, X.~Guo, W.~He, Y.~Ding, B.~Hong, M.~Zhang, J.~Wang, S.~Jin,
  E.~Zhou, R.~Zheng, X.~Fan, X.~Wang, L.~Xiong, Y.~Zhou, W.~Wang, C.~Jiang,
  Y.~Zou, X.~Liu, Z.~Yin, S.~Dou, R.~Weng, W.~Qin, Y.~Zheng, X.~Qiu, X.~Huang,
  Q.~Zhang, T.~Gui, The rise and potential of large language model based
  agents: a survey, Science China Information Sciences 68~(2) (2025) 121101.
\newblock \href {https://doi.org/10.1007/s11432-024-4222-0}
  {\path{doi:10.1007/s11432-024-4222-0}}.

\bibitem{RN41}
A.~Dorri, S.~S. Kanhere, R.~Jurdak, Multi-agent systems: A survey, IEEE Access
  6 (2018) 28573--28593.
\newblock \href {https://doi.org/10.1109/ACCESS.2018.2831228}
  {\path{doi:10.1109/ACCESS.2018.2831228}}.

\bibitem{RN42}
D.~Silver, A.~Huang, C.~J. Maddison, A.~Guez, L.~Sifre, G.~van~den Driessche,
  J.~Schrittwieser, I.~Antonoglou, V.~Panneershelvam, M.~Lanctot, S.~Dieleman,
  D.~Grewe, J.~Nham, N.~Kalchbrenner, I.~Sutskever, T.~Lillicrap, M.~Leach,
  K.~Kavukcuoglu, T.~Graepel, D.~Hassabis, Mastering the game of go with deep
  neural networks and tree search, Nature 529~(7587) (2016) 484--489.
\newblock \href {https://doi.org/10.1038/nature16961}
  {\path{doi:10.1038/nature16961}}.

\bibitem{RN12}
L.~WANG, A survey on large language model based autonomous agents, Front.
  Comput. Sci (2024).
\newblock \href {https://doi.org/10.1007/s11704-024-40231-1}
  {\path{doi:10.1007/s11704-024-40231-1}}.

\bibitem{RN79}
{Gartner, Inc.}, Top strategic technology trends for 2025,
  \url{https://www.gartner.com/en/articles/top-technology-trends-2025}
  (Accessed: 2025-03-24) (2024).

\bibitem{RN43}
OpenAI, Practices for governing agentic ai systems,
  \url{https://cdn.openai.com/papers/practices-for-governing-agentic-ai-systems.pdf}
  (Accessed: 2025-03-24) (2024).

\bibitem{RN67}
Y.~Wan, Empowering llms by hybrid retrieval-augmented generation for
  domain-centric q\&a in smart manufacturing, Advanced Engineering Informatics
  (2025).
\newblock \href {https://doi.org/10.1016/j.aei.2025.103212}
  {\path{doi:10.1016/j.aei.2025.103212}}.

\bibitem{RN68}
Q.~Xu, F.~Qiu, G.~Zhou, C.~Zhang, K.~Ding, F.~Chang, F.~Lu, Y.~Yu, D.~Ma,
  J.~Liu, A large language model-enabled machining process knowledge graph
  construction method for intelligent process planning, Advanced Engineering
  Informatics 65 (2025).
\newblock \href {https://doi.org/10.1016/j.aei.2025.103244}
  {\path{doi:10.1016/j.aei.2025.103244}}.

\bibitem{RN49}
C.-Y. Lin, Generative ai for intelligent manufacturing virtual assistants in
  the semiconductor industry, IEEE ROBOTICS AND AUTOMATION LETTERS. PREPRINT
  VERSION (2025).
\newblock \href {https://doi.org/10.1109/LRA.2025.3544506}
  {\path{doi:10.1109/LRA.2025.3544506}}.

\bibitem{RN47}
J.~Jeon, Y.~Sim, H.~Lee, C.~Han, D.~Yun, E.~Kim, S.~L. Nagendra, M.~B.~G. Jun,
  Y.~Kim, S.~W. Lee, J.~Lee, Chatcnc: Conversational machine monitoring via
  large language model and real-time data retrieval augmented generation,
  Journal of Manufacturing Systems 79 (2025) 504--514.
\newblock \href {https://doi.org/10.1016/j.jmsy.2025.01.018}
  {\path{doi:10.1016/j.jmsy.2025.01.018}}.

\bibitem{RN66}
J.~A.~H. ´Alvaro, An advanced retrieval-augmented generation system for
  manufacturing quality control, Advanced Engineering Informatics (2025).
\newblock \href {https://doi.org/10.1016/j.aei.2024.103007}
  {\path{doi:10.1016/j.aei.2024.103007}}.

\bibitem{RN70}
T.~Tu, S.~Azizi, D.~Driess, M.~Schaekermann, M.~Amin, P.-C. Chang, A.~Carroll,
  C.~Lau, R.~Tanno, I.~Ktena, A.~Palepu, B.~Mustafa, A.~Chowdhery, Y.~Liu,
  S.~Kornblith, D.~Fleet, P.~Mansfield, S.~Prakash, R.~Wong, S.~Virmani,
  C.~Semturs, S.~S. Mahdavi, B.~Green, E.~Dominowska, y.~Arcas Blaise~Aguera,
  J.~Barral, D.~Webster, S.~Corrado~Greg, Y.~Matias, K.~Singhal, P.~Florence,
  A.~Karthikesalingam, V.~Natarajan, Towards generalist biomedical ai, NEJM AI
  1~(3) (2024) AIoa2300138, doi: 10.1056/AIoa2300138.
\newblock \href {https://doi.org/10.1056/AIoa2300138}
  {\path{doi:10.1056/AIoa2300138}}.

\bibitem{RN71}
D.~Driess, F.~Xia, M.~S.~M. Sajjadi, C.~Lynch, A.~Chowdhery, B.~Ichter,
  A.~Wahid, J.~Tompson, Q.~H. Vuong, T.~Yu, W.~Huang, Y.~Chebotar, P.~Sermanet,
  D.~Duckworth, S.~Levine, V.~Vanhoucke, K.~Hausman, M.~Toussaint, K.~Greff,
  A.~Zeng, I.~Mordatch, P.~R. Florence, Palm-e: An embodied multimodal language
  model, in: International Conference on Machine Learning, 2023.

\end{thebibliography}
\bibliographystyle{elsarticle-num}







\end{document}